\documentclass{article}
\usepackage{ijcai24}

\usepackage{times}
\usepackage{soul}
\usepackage{url}
\usepackage[hidelinks]{hyperref}
\usepackage[utf8]{inputenc}
\usepackage[small]{caption}
\usepackage{graphicx}
\usepackage{amsmath}
\usepackage{amsthm}
\usepackage{booktabs}
\usepackage{algorithm}
\usepackage{algorithmic}
\usepackage[switch]{lineno}

\renewcommand{\phi}{\varphi}

\usepackage{times}  % DO NOT CHANGE THIS
\usepackage{helvet}  % DO NOT CHANGE THIS
\usepackage{courier}  % DO NOT CHANGE THIS
\usepackage{graphicx} % DO NOT CHANGE THIS
\usepackage{caption} % DO NOT CHANGE THIS AND DO NOT ADD ANY OPTIONS TO IT
\usepackage{algorithm}
\usepackage{algorithmic}

\newcommand{\nciteyear}[1]{[\citeyear{#1}]}

\usepackage{newfloat}
\newcommand{\Scal}{{\cal S}}

\newcommand{\U}{{\cal U}}

\newcommand{\cF}{{\cal F}}
\newcommand{\V}{{\cal V}}
\newcommand{\R}{{\cal R}}

\newcommand{\K}{{\cal K}}

\newcommand{\sat}{\models}

\newcommand{\commentout}[1]{}

\newcommand{\satt}{\models}
\newtheorem{theorem}{Theorem}
\newtheorem{lemma}{Lemma}
\newtheorem{definition}{Definition}
\newtheorem{proposition}{Proposition}
\newtheorem{example}{Example}

\newcommand{\lem}{\begin{lemma}}
\newcommand{\elem}{\end{lemma}}
\newcommand{\pro}{\begin{proposition}}
\newcommand{\epro}{\end{proposition}}

\newcommand{\dfn}{\begin{definition}}
\newcommand{\edfn}{\end{definition}}

\newcommand{\bbox}{\vrule height7pt width4pt depth1pt}
\newcommand{\eprf}{\bbox\vspace{0.1in}}

\newcommand{\xam}{\begin{example}}
\newcommand{\exam}{\end{example}}
\newcommand{\prf}{\noindent{\bf Proof.} }

\title{Explaining Image Classifiers}
\author{Hana Chockler$^1$\and
Joseph Y. Halpern$^2$\\
\affiliations
$^1$King's College, London, U.K\\
$^2$Cornell University, Ithaca, New York, USA\\
\emails
hana.chockler@kcl.ac.uk,
halpern@cs.cornell.edu
}
\begin{document}

\maketitle

\begin{abstract}
  % joe6: added an abstract
  We focus on explaining image classifiers, taking the work of
%joe11
  %  Mothilal et al. \nciteyear{MMTS21} (MMTS). We observe that,
    Mothilal et al. \nciteyear{MMTS21} (MMTS) as our point of
    departure. We observe that, 
although MMTS claim to be using the definition of explanation proposed
by Halpern \nciteyear{Hal48}, they do not 
quite do so. Roughly speaking, Halpern’s definition
has a necessity clause and a sufficiency
clause. MMTS replace the necessity clause by a requirement that, as we
show, implies it. Halpern’s definition also allows agents to restrict
the set of options considered.
%joe11
%We discuss the implications of MMTS’s simplification.
While these difference may seem minor, as we show, they can have a
nontrivial impact on explanations.
We also show that, essentially without change,
Halpern’s definition can handle two issues that have proved difficult
for other approaches: explanations of absence (when, for example, an
%hana4
image
classifier for tumors outputs “no tumor”) and explanations of rare
events (such as tumors).
\end{abstract}

\section{Introduction}
%joe6
%Black-box AI, and in particular Deep Neural Networks (DNNs) are now a
Black-box AI systems and, in particular Deep Neural Networks (DNNs), are now a
primary building block of many computer 
vision systems. DNNs are complex non-linear functions with algorithmically
%joe1: I don't know what it means for a coefficient to be "engineered"
%generated (and not engineered) coefficients. In contrast to
generated coefficients. In contrast to
%joe1
%traditionally engineered image processing pipelines it is difficult
%joe3
%traditional image processing pipelines, it is difficult
traditional image-processing pipelines, it is difficult  
to retrace how the 
%joe3
%pixel data are interpreted by the layers of the DNN.
pixel data are interpreted by the layers of a DNN.
%joe1
%This ``black box" nature of DNNs creates demand for
%joe3
%This ``black box" nature of DNNs creates demand for \emph{explanations}.
This ``black box" nature of DNNs creates a demand for \emph{explanations}.
A good explanation should answer the question 
``Why did the neural network classify the input the way it did?''
%joe1
%A good explanation can increase a user's confidence
By doing so, it can increase a user's confidence
in the result. Explanations are also useful for determining whether
%joe6: I don't think of vision systems as having "faults"
%there is a fault in the automated procedure:
the system is working well;
if the explanation does not make sense, it may indicate that 
%joe6
%the procedure is faulty.
%hana4
there is
a problem with the system.
%joe1: moved from below, with some rewriting, although we might want
%to cut it.

%joe1: added paragraph break
%It is less clear how to define what an explanation is, let alone a
%a\emph{good} explanation. There have been a number of definitions of
%joe4
%Unfortunately, It is not clear how to define what an explanation is, let alone 
Unfortunately, it is not clear how to define what an explanation is, let alone 
what a \emph{good} explanation is. There have been a number of definitions of
%joe3
%explanations over the years in various domains of computer
explanation given by researchers in various fields, particularly computer
%joe1
%science~\cite{CH97,Gar88,Pea88}, philosophy~\cite{Hem65} and
%joe3: added more philosophy references
%science~\cite{CH97,Gar88,Hal48,HP01a,Pea88}, philosophy~\cite{Hem65} and
%joe6
%science~\cite{CH97,Gardenfors1,Hal48,HP01a,Pea88} and
science \cite{CH97,Hal48,HP01a,Pearl} and
philosophy~\cite{Gardenfors1,Hempel65,Salmon70,Salmon89,Woodward14},
%joe3: oon is a philosopher, not a statistician!
%statistics~\cite{Sal89}.
%joe1*: added; we need to tell our story
%joe3
%There have been a number of attempts to provide explanations for the
and a number of attempts to provide explanations for the
%joe6: we need references here
%output of DNNs.  Halpern
%hana4
output of DNNs~\cite{lime,cam17,LL17,SCHK20,CKS21}
%joe7
%(see also an overview at~\cite{Mol19}).
(\cite{Mol19} provides an overview).
%joe4:
Here we focus on one particular definition of explanation, that 
%hana4
was
given
%joe6
%by \citeyear{Hal48}, which is in turn based on a
by Halpern \nciteyear{Hal48}, which is in turn based on a 
%joe6
%definition due to Halpern and Pearl \cite{HP01b}.   Mothilal et
definition due to Halpern and Pearl 
%hana4 changed the reference to explanations
\nciteyear{HP01a}.   
%\nciteyear{HP01b}.   
Mothilal et
al. \nciteyear{MMTS21} (MMTS from now on) already showed that this
definition could be usefully applied to better understand and evaluate
what MMTS called \emph{attribution-based} explanation approaches, such
as LIME \cite{lime} and SHAP \cite{LL17},
which provide a score or ranking over features, conveying the
%joe6
%(relative) importance of each feature to the model's output and
(relative) importance of each feature to the model's output, and
contrast them with what they called \emph{counterfactual-based}
%joe6
%approaches, such as DICE \cite{MMTS20} and that of Wachter et
%which generate examples that yield a different model
approaches, such as DICE \cite{MMTS20} and that of
%joe16
%Wachter, Mittelstadt, C. Russell \nciteyear{WMR17},
Wachter et al. \nciteyear{WMR17}, 
which generate examples that yield a different model
output with minimum changes in the input features.    

%joe4: added.  This is what I think we can salvage from what was written
%joe12:
%In this note, we take MMTS as our point of departure and focus on
In this paper, we take MMTS as our point of departure and focus on
explaining image classifiers. We first observe that, 
although MMTS claim to be using Halpern's definition, they do not 
%joe6
%so in two respects.  Roughly speaking, Halpern's definition (which we
%they replace they do not do so in two respects. Roughly speaking,
%Halpern’s definition (which we discuss in detail in Section 2) has a
%necessity clause and a sufficiency clause. In both cases, MMTS
%replace these clauses by ones that imply them;   
%for the necessity clause, they point this out explicitly.
%We discuss the implications of these simplifications.  We
quite do so.  Roughly speaking, Halpern's definition (which we
discuss in detail in Section~\ref{sec:background}) has a necessity clause and a
sufficiency clause.  MMTS replace the necessity clause by a
requirement that, as we show, implies it. Halpern's definition also
allows agents to restrict the set of options considered.
%joe11
%We discuss the implications of MMTS's simplification.
While these difference may seem minor, as we show, they can have a
nontrivial impact on explanations.
We
also show that, essentially without change, Halpern's definition can handle two
issues that have proved difficult for other approaches: explanations
of absence (when, for example, an classifier for tumors outputs ``no
tumor'') and explanations of rare events (again, a classifier for
tumors can be viewed as an example; a tumor is a relatively rare
event).  
%joe11
The upshot of this discussion is that, while the analysis of MMTS
shows that a simplification of Halpern's definition can go a long way
to helping us understand notions of explanation used in the
literature, we can go much further by using Halpern's actual
definition, while still retaining the benefits of the MMTS analysis.

%joe1*: I cut this; I think it distracts from our story
\commentout{
The recent increase in the number of machine learning applications and the advances in deep learning led to the need
for \textit{explainable AI}, which is advocated, among others, by
DARPA~\cite{DARPA} to promote understanding, trust, and adoption of
future autonomous systems based on learning algorithms (and, in
particular, image classification DNNs). Miller~\nciteyear{Mil19}
presents an overview of different types of explanations existing in
the literature and those expected by the human users of AI tools. 
He broadly describes an explanation as an answer to a why-question, 
which in our context would be \emph{``why did the neural network
classify the input the way it did?''}.
}
 
%joe4*: cut this; I don't think it's new anymore
\commentout{
Existing algorithms for generating explanations can be roughly divided
%joe1
%into \emph{white-box}, which rely on a specific internal structure of
into \emph{white-box} algorithms, which rely on the internal structure of
the network and use specific layers to extract explanations, and
%joe1
%\emph{black-box}, which are agnostic to the internal structure of the
\emph{black-box} algorithms, which are agnostic regarding the internal
structure of the 
network and use perturbations of the input and their effect on the
output to deduce the importance of different features of the input on
the output~\cite{RSB18}.  
%joe1
%In this paper we focus on black-box explanations of image
%classifiers.
%joe3
%In this paper we focus on black-box algorithms for explaining the
%output of image classifiers. 
In this paper, we focus on black-box algorithms for explaining the
output of image classifiers.% 
%hana1
%joe3
%\footnote{Almost all tools we mention in the paper can provide
\footnote{Almost all the tools we mention in the paper can provide
explanations 
%joe3
%of classifiers for other input formats, and our conclusions hold for
of classifiers for other types of input; our conclusions hold for
these other 
types as well.}

%joe1
%The black-box explanation tools can be roughly divided into those that
Black-box algorithms can be roughly divided into those that
%joe3: I don't like the word "small" in this context.  How about
%basic?  I made the change globally
%provide small
provide what we call \emph{basic}
%hana1 I now think that "minimal" is an orthogonal feature
%(close to \emph{minimal}) 
explanations and those that provide \emph{robust}
%joe1* I understand that what say below is the case for the DeepCover
%algorithm that you developed.  Is it really the case for all the
%other approaches?  It certainly doesn't seem to be the case for LIME
%or SHAP.  If it is, you need to explain how.
%explanations. Roughly speaking, a basic explanation is a part of the
explanations. Roughly speaking, a \emph{basic explanation} is a part of the
input image 
that is sufficient for the image classifier to give the same output as
%joe1
%for the whole input, when the rest of the input image is obscured
for the whole input if the rest of the input image is obscured
(replaced 
with a neutral color). 
%hana1 brought the definition of a robust explanation up and minimal after that
A \emph{robust explanation}, in contrast, is a part of the input image that is
sufficient for the image classifier to give the 
%joe1
%same output regardless of the values of other parts of the input; that
%is, a robust explanation of, say, an image classified as a dog, is a
same output regardless of the pixel values of other parts of the 
%hana1
%joe3: is this what you meant?
%input. In fact, instead of completely freeing the pixel values of
%other parts of the input, 
%the existing tools for providing robust explanations superimpose another image
%over the rest of the pixels. That is, a robust explanation of, say, an
input. (For computational reasons, rather than considering all possible
pixel values, existing tools for providing robust
explanations take an explanation to be robust if it is sufficient for
the classifier to give the same output regardless of how the remaining
pixels are replaced by other images.  For example, a robust
explanation of an 
image classified as 
a dog is a part of the image such that all images with this part fixed
and other parts 
%joe3
%replaced by another image superimposed over them.
replaced other images still result in the classification ``dog''.)
%input; that
%is, a robust explanation of, say, an image classified as a dog is a
%part of the 
%%joe1*: I don't understand what it means to overlay one image on
%%another in this context.  Could you clarify?
%%image that is sufficient to be classified as a dog even when overlayed
%image that is sufficient to be classified as a dog even when overlaid
%on any other image, for example, of a person.
%hana1 re-defined minimal 
%joe3: We don't use "minimal" here.  There's no reason to define it
%here, and it's a distraction
%An explanation is \emph{minimal} if no strict subset of it is in
%itself an explanation according to the same definition. 
%%joe1
%``Minimal'' in this context means that no strict
%subset of an explanation in itself an explanation. 
%joe1: moved from below and rewrote
It is easy to see that any robust explanation contains a 
%hana1
basic
%minimal
explanation.    What counts as a robust explanation will in general depend on the
size and diversity of the dataset.

%joe1: I think this is a distraction
%Minimal (or, more precisely, close to minimal) explanations have been more
%popular, as they are generally considered to be more helpful to the
%user than  
%large ones.
%joe1*

%joe1*: We may want to move this.
%There are a number of tools that provide close to minimal
%explanations, which we briefly survey below.
We briefly review the main 
%hana1
black-box
approaches that have
been used to provide 
%hana
explanations of image classifiers.
%explanations.
The first three provide
%hana1
basic
%minimal 
%
explanations; the last one provides robust explanations.%
%hana1
%joe3: let's make this a footnote; it's a bit of distraction
\footnote{
All tools in the list below segment the input image to \emph{superpixels}, which are basic areas of the image, and treat these
superpixels as atomic elements of the image. Essentially, the division to superpixels instead of the individual pixels is done purely
for scalability reasons. In the rest of the paper, everywhere we refer to pixels, the reader should keep in mind that real applications
deal with superpixels instead.}  
\begin{itemize}
  \item
    SHAP (Shapley Additive exPlanations)~\cite{LL17}
%joe3*: what does identifying inputs that are similar to the given
    %input have to do with anything?  Rewrote; please check.
    %    identifies inputs
%that are similar to the input for
%which the output is to be explained, according to some distance
%measure, and uses Shapley value~\cite{Shapley1953} to rank the
%%hana1
%pixels.
    %elements of the input.
    %    Roughly speaking, the Shapley value of a set of pixels can be viewed
%as the measuring the contribution of that set to the final output
    %joe1: added
computes the \emph{Shapley value} of each set of pixels, 
which can be viewed
as a measurement of the contribution of that set to the final output
determination.   
%joe1
%A subset of
%the elements with the highest Shapley values are presented as an explanation.
The subset of 
%hana1
pixels
%elements 
with the highest Shapley values is then
presented as the explanation.
%joe1: Above you said all black-box method use perturbation.  But now
%you're calling 
%In contrast, perturbation-based explanation approaches explore the
%input space by introducing interventions (perturbations) 
%in search of an explanation. 
\item
%joe3
%  Given a particular input, LIME~\cite{lime} samples the neighborhood of
  Given an  input, LIME~\cite{lime} samples the neighborhood of
  this input and uses local perturbations, 
%hana1
%joe3*: I don't know what it means to hide a pixel, nor do I have any
%idea how doing so could be used to rank the pixels.  Can you clarify?
  namely, hiding pixels, 
to rank the 
%hana1
pixels
%elements 
of the
%joe1*: Hana, you have to explain how LIME uses perturbation to ran the
%elements of the input!  I certainly have no idea, based on what you wrote.
input; an explanation is, again, a subset of the 
%hana1
pixels
%elements 
with the
highest rank.
\item
%joe1
%  DeepCover~\cite{SCHK20,CKS21}, explicitly defines an explanation as a
%joe3
%  DeepCover~\cite{SCHK20,CKS21} explicitly defines an explanation as a
%hana2 added the most updated citation and changed the name to ReX
  ReX~\cite{CKK23} and its previous version DeepCover~\cite{SCHK20,CKS21} take an explanation to be a
  minimal part of the input that is sufficient for obtaining the same classification as the original input if the rest of the input is obscured.
  ReX computes an approximation of an explanation
by first ranking 
%hana1
pixels
%parts of the input 
according to their approximate
%joe1: how is the approximate degree of responsibility computed?  I
%wouldn't mind cutting this sentence altogether, but if we're going to
%include it, we have to say a little more.
%hana1*
%it is done according to the definition: by finding a basicest subset of pixels so that changing
%all of these results in a change in the classification
degree of responsibility in the classification, and then constructing 
an explanation by greedily adding the ranked 
%hana1
pixels
%parts 
to an empty input
until the classification is the same as for the original input. 
\item  
%joe1*: Hana, you have to say something about how Anchors works!
%joe3
  %  Anchors~\cite{RSG18} finds \emph{robust} explanations
    Anchors~\cite{RSG18} finds robust explanations
%hana1
    by fixing some pixels to their original values in the input image and
%joe3
%    superimposing another image from the dataset over the rest of the pixels.
replacing the remaining pixels by another image from the dataset.
    %joe3: I'm afraid I still don't understand this description.    How
%does Anchor decide what to return? Is the following correct?
Anchors returns a part of the image
that is sufficient to guarantee the same classification regardless of
how this replacement is done.
\end{itemize}
  %joe1: we've already defined robust explanation
%  , that is, parts of an input image that are sufficient for getting
%  the same classification 
%as the whole input when overlayed on \emph{any} other
%image in the dataset. It is easy to see that any robust explanation
%contains a minimal explanation.  
% Robust explanations depend on the size and diversity of the
%dataset. Theoretically, although this is not implemented in any tool,
%we can also define the most robust explanation as a part of the input
%image that results in the same classification as the original input
%regardless of the values of the 
%rest of the input.
%\end{itemize}

But in what sense are the 
%hana1
basic
%minimal 
or robust explanations that these
systems provide really explanations?
To understand this issue better, we consider what would count as an
explanation using the definition of explanation given by Halpern
\nciteyear{Hal48}, which is in turn based on a 
%joe8
%definition due to Halpern and Pearl \cite{HP01b}.  We are then able
definition due to Halpern and Pearl \nciteyear{HP01b}.  We are then able
to relate this definition to what is actually done by the various
algorithms, and see how close they come to giving a true explanation.
%joe1*: It would be good if we can relate some of the other approaches
%to explanation too!
We believe that this exercise both gives deeper insight into what
these tools are doing, and could lead to improved tools that will
prove yet more useful for users.  
}
%joe4: \end{commentout}

The rest of the paper is organized as follows:  In
Section~\ref{sec:background}, we review the relevant definitions of
%hana3
%joe6
%causal models and explanations, in Section~\ref{sec:imagexp} we
causal models and explanations; in Section~\ref{sec:imagexp}, we
discuss explanations 
of image classifiers and show their relation to explanations in actual
%joe6
%causality, and in Section~\ref{sec:beyond} we discuss explanations of
causality; and in Section~\ref{sec:beyond}, we discuss explanations of
absence and of rare events. 

\section{Causal Models and Relevant Definitions}\label{sec:background} 

In this section, we review the definition of causal
models introduced by Halpern and Pearl \nciteyear{HP01b} and
%joe1
%several definitions of causes and explanations given by Halpern.
%joe6
%relevant definitions of causes and explanations given by Halpern \cite{Hal48}.
relevant definitions of causes and explanations given by Halpern
\nciteyear{Hal48}. 
The material in this section is largely taken from \cite{Hal48}.

We assume that the world is described in terms of 
variables and their values.  
Some variables may have a causal influence on others. This
influence is modeled by a set of {\em structural equations}.
It is conceptually useful to split the variables into two
sets: the {\em exogenous\/} variables, whose values are
determined by 
factors outside the model, and the
{\em endogenous\/} variables, whose values are ultimately determined by
the exogenous variables.  
The structural equations
describe how these values are 
determined.

Formally, a \emph{causal model} $M$
is a pair $(\Scal, \cF)$, where $\Scal$ is a \emph{signature}, which explicitly
lists the endogenous and exogenous variables  and characterizes
their possible values, and $\cF$ defines a set of \emph{(modifiable)
structural equations}, relating the values of the variables.  
A signature $\Scal$ is a tuple $(\U,\V,\R)$, where $\U$ is a set of
exogenous variables, $\V$ is a set 
of endogenous variables, and $\R$ associates with every variable $Y \in 
\U \cup \V$ a nonempty set $\R(Y)$ of possible values for 
%joe6: inside parens, you should use "i.e." and "e.g."
%$Y$ (that is, the set of values over which $Y$ {\em ranges}).
$Y$ (i.e., the set of values over which $Y$ {\em ranges}).  
For simplicity, we assume here that $\V$ is finite, as is $\R(Y)$ for
every endogenous variable $Y \in \V$.
$\cF$ associates with each endogenous variable $X \in \V$ a
function denoted $F_X$
(i.e., $F_X = \cF(X)$)
such that $F_X: (\times_{U \in \U} \R(U))
\times (\times_{Y \in \V - \{X\}} \R(Y)) \rightarrow \R(X)$.
This mathematical notation just makes precise the fact that 
$F_X$ determines the value of $X$,
given the values of all the other variables in $\U \cup \V$.
If there is one exogenous variable $U$ and three endogenous
variables, $X$, $Y$, and $Z$, then $F_X$ defines the values of $X$ in
terms of the values of $Y$, $Z$, and $U$.  For example, we might have 
$F_X(u,y,z) = u+y$, which is usually written as
$X = U+Y$.   Thus, if $Y = 3$ and $U = 2$, then
$X=5$, regardless of how $Z$ is set.%
\footnote{The fact that $X$ is assigned  $U+Y$ (i.e., the value
of $X$ is the sum of the values of $U$ and $Y$) does not imply
that $Y$ is assigned $X-U$; that is, $F_Y(U,X,Z) = X-U$ does not
necessarily hold.}  

The structural equations define what happens in the presence of external
interventions. 
Setting the value of some variable $X$ to $x$ in a causal
model $M = (\Scal,\cF)$ results in a new causal model, denoted
$M_{X\gets x}$, which is identical to $M$, except that the
equation for $X$ in $\cF$ is replaced by $X = x$.

%joe7
%We can also consider \emph{probabilistic causal models}
We can also consider \emph{probabilistic causal models};
%joe1
%if we want to talk about the probability of causality (and, for our
%purposes, the probability of discrimination).  A probabilistic causal
%model is a tuple
%$M=(\Scal,\cF,\Pr)$, where $(\Scal,\cF)$ is a causal model, and $\Pr$ is
%a probability on contexts.
%hana4
%joe7
%as pairs $(M,\Pr)$,
these are pairs $(M,\Pr)$, 
%is a  pair $(M,\Pr)$, 
where $M$ is a causal model and $\Pr$ is a
probability on the contexts in $M$.

The dependencies between variables in a causal model $M = ((\U,\V,\R),\cF)$
can be described using a {\em causal network}\index{causal
  network} (or \emph{causal graph}),
whose nodes are labeled by the endogenous and exogenous variables in
$M$, with one node for each variable in $\U \cup
\V$.  The roots of the graph are (labeled by)
the exogenous variables.  There is a directed edge from  variable $X$
to $Y$ if $Y$ \emph{depends on} $X$; this is the case
if there is some setting of all the variables in 
$\U \cup \V$ other than $X$ and $Y$ such that varying the value of
$X$ in that setting results in a variation in the value of $Y$; that
is, there is 
a setting $\vec{z}$ of the variables other than $X$ and $Y$ and values
$x$ and $x'$ of $X$ such that
$F_Y(x,\vec{z}) \ne F_Y(x',\vec{z})$.

A causal model  $M$ is \emph{recursive} (or \emph{acyclic})
if its causal graph is acyclic.
It should be clear that if $M$ is an acyclic  causal model,
then given a \emph{context}, that is, a setting $\vec{u}$ for the
exogenous variables in $\U$, the values of all the other variables are
determined (i.e., there is a unique solution to all the equations).
In this paper, following the literature, we restrict to recursive models.

\commentout{
Given a signature $\Scal= (\U,\V,\R)$, a \emph{primitive event} is a
formula of the form $X = x$, for  $X \in \V$ and $x \in \R(X)$.  
A {\em causal formula (over $\Scal$)\/} is one of the form
$[Y_1 \gets y_1, \ldots, Y_k \gets y_k] \varphi$,
where
$\varphi$ is a Boolean
combination of primitive events,
$Y_1, \ldots, Y_k$ are distinct variables in $\V$, and
$y_i \in \R(Y_i)$.
Such a formula is abbreviated
as $[\vec{Y} \gets \vec{y}]\varphi$.
The special
case where $k=0$
is abbreviated as
$\varphi$.
Intuitively,
$[Y_1 \gets y_1, \ldots, Y_k \gets y_k] \varphi$ says that
$\varphi$ would hold if
$Y_i$ were set to $y_i$, for $i = 1,\ldots,k$.
}

We call a pair $(M,\vec{u})$ consisting of a causal model $M$ and a
context $\vec{u}$ a \emph{(causal) setting}.
%joe6: once is enough :-)
%We call a pair $(M,\vec{u})$ consisting of a causal model $M$ and a
%context $\vec{u}$ a \emph{(causal) setting}.
A causal formula $\psi$ is true or false in a setting.
We write $(M,\vec{u}) \satt \psi$  if
the causal formula $\psi$ is true in
the setting $(M,\vec{u})$.
The $\satt$ relation is defined inductively.
$(M,\vec{u}) \satt X = x$ if
the variable $X$ has value $x$
in the unique (since we are dealing with acyclic models) solution
to the equations in
%joe6: 
%$M$ in context $\vec{u}$ (that is, the
$M$ in context $\vec{u}$ (i.e., the
unique vector of values for the exogenous variables that simultaneously satisfies all
equations 
in $M$ 
with the variables in $\U$ set to $\vec{u}$).
Finally, 
$(M,\vec{u}) \satt [\vec{Y} \gets \vec{y}]\varphi$ if 
$(M_{\vec{Y} = \vec{y}},\vec{u}) \satt \varphi$,
where $M_{\vec{Y}\gets \vec{y}}$ is the causal model that is identical
to $M$, except that the 
equations for variables in $\vec{Y}$ in $\cF$ are replaced by $Y = y$
for each $Y \in \vec{Y}$ and its corresponding 
value $y \in \vec{y}$.

A standard use of causal models is to define \emph{actual causation}: that is, 
what it means for some particular event that occurred to cause 
 another particular event. 
There have been a number of definitions of actual causation given
for acyclic models
(e.g., \cite{beckers21c,GW07,Hall07,HP01b,Hal48,hitchcock:99,Hitchcock07,Weslake11,Woodward03}).
%joe1
%In this paper, we focus on what \cite{Hal48} calls the \emph{modified}
In this paper, we focus on what has become known as the \emph{modified}
%joe6
%Halpern-Pearl definition and on other, related, definitions introduced
%in ~\cite{Hal48}. 
Halpern-Pearl definition and some related definitions introduced
by Halpern ~\nciteyear{Hal48}. 
We briefly review the relevant definitions below
%joe6
%(see \cite{Hal48} for more intuition and motivation.)
(see \cite{Hal48} for more intuition and motivation). 

The events that can be causes are arbitrary conjunctions of primitive
events (formulas of the form $X=x$); 
the events that can be caused are arbitrary Boolean combinations of primitive events.  
an arbitrary formula $\phi$. 
%joe1: not in this paper
%we restrict ourselves to defining what it
%means for $\vec{X}=\vec{x}$ \emph{rather than $\vec{X} = \vec{x}'$} to be a
%cause of $O=o$ rather than $O=o'$.

\dfn\label{def:AC}[Actual cause]
$\vec{X} = \vec{x}$ is 
an \emph{actual cause} of $\varphi$ in $(M,\vec{u})$ if the
following three conditions hold: 
\begin{description}
\item[{\rm AC1.}]\label{ac1} $(M,\vec{u}) \models (\vec{X} = \vec{x})$ and $(M,\vec{u}) \models \varphi$. 
\item[{\rm AC2.}] There is a
  %joe3*
%    set $\vec{W}$ of variables in $\V$
  a setting $\vec{x}'$ of the variables in $\vec{X}$, a 
(possibly empty)  set $\vec{W}$ of variables in $\V - \vec{X}'$,
and a setting $\vec{w}$ of the variables in $\vec{W}$
such that $(M,\vec{u}) \models \vec{W} = \vec{w}$ and
$(M,\vec{u}) \models [\vec{X} \gets \vec{x}', \vec{W} \gets
%joe3
  %  \vec{w}]\neg{\varphi}$.
    \vec{w}]\neg{\varphi}$, and moreover
\item[{\rm AC3.}] \label{ac3}\index{AC3}  
  $\vec{X}$ is minimal; there is no strict subset $\vec{X}'$ of
%joe3
%  $\vec{X}$ such that $\vec{X}' = \vec{x}''$ satisfies
  $\vec{X}$ such that $\vec{X}' = \vec{x}''$ can replace $\vec{X} =
  \vec{x}'$ in 
  AC2, where $\vec{x}''$ is the restriction of
%joe3*
%$\vec{x}$ to the variables in $\vec{X}'$.
$\vec{x}'$ to the variables in $\vec{X}'$.
\end{description}
\edfn
\noindent AC1 just says that $\vec{X}=\vec{x}$ cannot
be considered a cause of $\varphi$ unless both $\vec{X} = \vec{x}$ and
$\varphi$ actually holds.  AC3 is a minimality condition, which says
that a cause has no irrelevant conjuncts.  
AC2 extends the standard
%joe6
%but-for condition ($\vec{X}=\vec{x}$ s a cause of $\varphi$ if, had 
but-for condition ($\vec{X}=\vec{x}$ is a cause of $\varphi$ if, had 
$\vec{X}$ been $\vec{x}'$, $\varphi$ would have been false)
by allowing us to apply it while keeping 
%hana4
some variables fixed
%fixed some variables 
to the value that they had in the actual 
setting $(M,\vec{u})$.
%joe6: redundant
%In the special case that $\vec{W} = \emptyset$, we get the standard
%but-for definition of causality:  if $\vec{X} = \vec{x}$ had not
%occurred$\varphi$ would not have occurred.
In the special case that $\vec{W} = \emptyset$, we get the 
but-for definition.

To define explanation, we need the notion of \emph{sufficient cause}
in addition to that of actual cause.
%The definition below captures a different feature of causality: how
%sensitive the causality ascription is to changes in various other
%factors.

\dfn\label{def:SC}[Sufficient cause]
$\vec{X} = \vec{x}$ is 
a \emph{sufficient cause} of $\varphi$ in $(M,\vec{u})$
if the following four conditions hold: 
\begin{description}
\item[{\rm SC1.}] \label{sc1} $(M,\vec{u}) \models (\vec{X} = \vec{x})$ and $(M,\vec{u}) \models \varphi$. 
\item[{\rm SC2.}] \label{sc2} Some conjunct of $\vec{X} = \vec{x}$ is part of 
%hana10
an actual cause
%a cause 
of $\varphi$ in $(M,\vec{u})$. 
More precisely, there exists a conjunct $X=x$ of $\vec{X} = \vec{x}$ and another (possibly empty) conjunction $\vec{Y} = \vec{y}$ 
such that $X=x \wedge \vec{Y} = \vec{y}$ is an actual cause of $\varphi$ in $(M,\vec{u})$.  
\item[{\rm SC3.}] \label{sc3} $(M,\vec{u}') \models [\vec{X} =
%joe1
  %  \vec{x}]\varphi$  for all contexts $\vec{u}'$.
    \vec{x}]\varphi$  for all contexts $\vec{u}' \in 
%hana5 changed to \U because this is how it is referred to later
%joe8: corrected
  %  \U$.
\R(\U)$.  
 %   U$. 
\item[{\rm SC4.}] \label{sc4} $\vec{X}$ is minimal; there is no strict subset $\vec{X}'$ of $\vec{X}$ such that $\vec{X}' = \vec{x}'$ satisfies conditions
SC1, SC2, and SC3, where $\vec{x}'$ is the restriction of $\vec{x}$ to the variables in $\vec{X}'$.
\end{description}
\edfn
%joe1: expanded
\noindent
%joe11
Note that this definition of sufficient cause (which is taken from
\cite{Hal48}) is quite different from that in \cite{HP01b}.  
Like the necessity clause used by MMTS, the definition of \cite{HP01b} 
requires only that some
subset of $\vec{X} = \vec{x}$ be an actual cause of $\phi$, without
%joe14
%allowing the subset to be extend by another conjunction $\vec{Y} =
allowing the subset to be extended by another conjunction $\vec{Y} =
\vec{y}$ (and uses a different definition of actual cause---that of
\cite{HP01a}), but (somewhat in the spirit of SC3) requires that this
necessity condition hold in all contexts.  It has no exact analogue of
SC3 at all.
An example might help clarify the definition.

%hana4* I don't understand the example. how would you express it with the same model M? seems that the equations are different. but the definition of context
%only includes the values of exogenous variables, not the equations.
%joe7: the context is also a parent of FB (so can influence the
%equation for FB).  This is explained below, where I discuss the
%equation for FB.  I'm not sure what else to add (although I certainly
%don't object to adding something if it would help clarify things).
Suppose that we have a dry
%joe6
%forest, and three arsonists.  There are three contexts: in $u_1$,
%joe8: undid previous change
%forest and three arsonists.  $\U$ consists of three contexts: in $u_1$,
forest and three arsonists.  There are three contexts: in $u_1$,
it takes just one dropped match to burn the forest down, but 
arsonist 1 and arsonist 2 drop matches; in $u_2$, it takes just one dropped
match to burn the forest down in and all three arsonists drop a match;
finally, in $u_3$, it takes two 
%joe6
%droped matches to burns the forest down, and arsonists 1 and 3 drop matches.
dropped matches to burn the forest down, and arsonists 1 and 3 drop matches.
To model this, we have binary variables $ML_1$, $ML_2$, and $ML_3$,
%joe6
%denoting which arsonist drops a match, and $FB$ denoting whether the
denoting which arsonist drops a match, and $FB$ denoting, whether the
forest burns down.  (Note that we use the structural equation for $FB$
to capture these differences; for example, if $x_i$ is the value of
$ML_i$, then $F_{FB}(x_1,x_2,x_3,u_1) = 1$ iff at least one of $x_1$,
%joe6
%$x_2$, and $x_3$ are 1, and $F_{FB}(x_1,x_2,x_3,u_3) = 1$ iff at least
%one of $x_1$, $x_2$, and $x_3$ are 1.)
$x_2$, and $x_3$ is 1, and $F_{FB}(x_1,x_2,x_3,u_3) = 1$ iff at least
two of $x_1$, $x_2$, and $x_3$ are 1.)
We claim that $ML_1 = 1 \land ML_2 = 1$ is a sufficient cause of $FB =
%joe6: have defined "relative to U
%1$ relative to $U$ in $(M,u_1)$ and $(M,u_2)$, but not $(M,u_3)$.  To
1$ in $(M,u_1)$ and $(M,u_2)$, but not $(M,u_3)$.  To
see that it is not a sufficient cause in $(M,u_3)$, note that SC1 does
not hold: arsonist 2 does not drop a match.   It is also easy to see that
$(M,u_i) \sat [ML_1 = 1 \land ML_2 = 1](FB=1)$ for $i = 1, 2, 3$, so
SC3 holds.  
Now in  $(M,u_1)$, $ML_1 \land ML_2 = 1$ is an actual cause of $FB=1$
(since the values of both $ML_1$ and $ML_2$ have to be changed in
order to change the value of $FB$). Similarly, in $(M,u_2)$, $ML_1
\land ML_2 = 1 \land ML_3 = 1$ is an actual cause of $FB=1$ (so we get
SC2 by taking $Y = ML_3$).  Thus, SC2 holds in both $(M,u_1)$ and $(M,u_2)$.

%hana9
%joe14: SC2 looks so different from SC3.  I don't see a priori why SC3
%should imply SC2, because SC2 requires \neg \phi to hold sometimes.
%It might seem that sufficiency is a stronger condition than actual
%causality. However, in the general case, SC1, SC3, and SC4 do not
%imply SC2, as the following example demonstrates.
While SC3  is typically taken to be the core of the sufficiency
requirement, to show causality, we also need SC2, since it requires
that $\neg \phi$ hold for some setting of the variables.  We might hope
that if there were a setting where $\phi$ was false, SC3 would imply
SC2.  As the following example shows, this is not the case in general.

\begin{example}\label{ex:suffic}
  Consider a causal model $M$ with three
  %joe14
  binary
  variables $A$, $B$, and $C$, and
the structural equations 
$A=B$ and $C = A \vee (\neg{A} \wedge B)$. Let $\vec{u}$ be a context
%joe14
%in which all variables are set to $1$, and let $\varphi = C$. 
%Now, $A=1$ satisfies SC1, SC3, and SC4 for $\varphi$, however it does
%but it does not satisfy SC2. Indeed, $A$ by itself is not an actual 
%cause of $C$, as $B$ holds as well, and $A=1$ is not a part of an actual
%cause, as $B=1$ is already an actual cause of $C$.
in which all variables are set to $1$, and let $\varphi$ be the
formula $C=1$. 
In context $\vec{u}$, $A=1$ satisfies SC1, SC3, and SC4 for $\varphi$.
There is also some setting of the variables for which $C=0:$ 
$(M,\vec{u}) \sat [A \gets 0, B \gets 0](C=0)$.  However,
$A=1$ does not satisfy SC2. Indeed, $A=1$ by itself is not an actual 
cause of $C=1$, as $B$=1 holds as well, nor is $A=1$ part of an actual
cause, as $B=1$ is already an actual cause of $C=1$. 
\end{example}
On the other hand, if there are no dependencies between the variables 
and 
%joe14:  I don't see any reason why it should be the case in general
%the case in explanations of image classifiers, as we
%argue in Section~\ref{sec:imagexp} (this is also the assumption in MMTS),
some other assumptions made by MMTS in their analysis of image
classifiers hold,
%joe14
%then SC1, SC3, and SC4 imply SC2, as proved in the following theorem.
then, roughly speaking, SC1, SC3, and SC4 do imply SC2.  To make this
precise, we need  two definitions.

\begin{definition}
The variables in a set $\vec{X}$ of endogenous variables are \emph{causally
independent} in a causal model $M$ if, for all contexts $\vec{u}$,
all strict subsets  $\vec{Y}$ of $\vec{X}$, and all $Z \in \vec{X} -
\vec{Y}$, $(M,\vec{u}) \sat Z=z$ iff $(M,\vec{u}) \sat [\vec{Y} \gets
  \vec{y}](Z=z)$.
\end{definition}
Intuitively, the variables in $\vec{X}$ are causally independent if
setting the values of some subset of the variables in $\vec{X}$ has no
impact on the values of the other variables in $\vec{X}$.  

\begin{definition}
$\vec{X}$ is \emph{determined by the context} if for
each setting $\vec{x}$ of $\vec{X}$, there is a context
$\vec{u}_{\vec{x}}$ such that $(M,\vec{u}_{\vec{x}}) \sat X=x$.
  \end{definition}

\begin{theorem}\label{thm:nodepend}
%joe14: this is not quite what we need
  %  Given a causal model $M$, we say that its variables $\V$ are
%independent of each other if for each $V \in \V$, there exists a 
%unique $U \in \U$ such that $V = f(U)$ for some function $f$.
Given a set $\vec{X}'$ of endogenous variables in  a causal model $M$
such that (a) the variables in $\vec{X}'$ are causally independent,
(b) $\vec{X}'$ is determined by the context, (c) $\vec{X}'$ includes
all the parents of the 
variables in $\phi$, (d) there is some setting $\vec{x}'$ of the
variables in $\vec{X}'$ that 
makes $\phi$ false in context $\vec{u}$ (i.e., $(M,\vec{u}) \sat
[\vec{X}' \gets \vec{x}']\neg \phi$), and (e) $\vec{X} \subseteq \vec{X}'$,
then $\vec{X} = \vec{x}$ is a sufficient cause of $\varphi$ in $(M,\vec{u})$ iff
it satisfies SC1, SC3, and SC4 (i.e., SC2 follows).
\end{theorem}
%joe14*: added proof
\prf  By assumption, there is some setting $\vec{x}'$  such that
$\vec{X}'$ that makes $\phi$ false in context $\vec{u}$ (i.e.,
$(M,\vec{u}) \sat [\vec{X}' \gets \vec{x}']\neg \phi$).  Choose
$\vec{x}'$ to be such a setting that differs minimally from the values that
variables get in $\vec{u}$; that is, the set of variables
$Y \in \vec{X}$ such that $(M,\vec{u}) \sat Y = y$, and the value of
$Y$ in $\vec{x}'$ is different from $Y$ is minimal.
Let $\vec{Y}$ be the set of variables in this minimal set.
Let $\vec{u}_{\vec{x}'}$ be a context such that
$(M,\vec{u}_{\vec{x}'}) \sat \vec{X}' = \vec{x}'$; by assumption, such
a context exists.  Since $\vec{X}'$ includes all the parents of
the variables in $\phi$, we must have 
$(M,\vec{u}_{\vec{x}'}) \sat [\vec{X}' \gets \vec{x}']\neg \phi$.
Since $(M,\vec{u}_{\vec{x}'}) \sat \vec{X}' \gets \vec{x}'$, it
follows that $(M,\vec{u}_{\vec{x}'}) \sat \neg \phi$.  

$\vec{Y}$ must contain a variable in $\vec{X}$.  For suppose not. By SC1,
$(M,\vec{u}) \sat \vec{X} = \vec{x}$, so if $Y$ does not
contain a variable in $X$, the values that the variables in $\vec{X}$
get in the setting $\vec{X}' = \vec{x}'$ is the same as their value in
$(M,\vec{u})$.  Thus,  $(M,\vec{u}_{\vec{x}'}) \sat \vec{X} = \vec{x}$.
By SC3, $(M,\vec{u}_{\vec{x}'}) \sat [\vec{X} = \vec{x}]\phi$, from
which it would follow that $(M,\vec{u}_{\vec{x}'}) \sat \phi$, a
contradiction.

Let $\vec{y}'$ be the restriction of $\vec{x}'$ to the variables in
$\vec{Y}$, and let $\vec{y}$ be the values of the variables in
$\vec{Y}$ in $(M,\vec{u})$, that is $(M,\vec{u}) \sat \vec{Y} =
\vec{y}$.  We claim that $\vec{Y} = \vec{y}$ is a cause of $\phi$ in
$(M,\vec{u})$.  By assumption, $(M,\vec{u}) \sat \vec{Y} = \vec{y}$; by
SC1, $(M,\vec{u}) \sat \phi$.  Thus, AC1 holds.  By construction,
$(M,\vec{u}) \sat [\vec{Y} \gets \vec{y}']\neg \phi$, so AC2 holds
  (with $\vec{W} = \emptyset$).  Finally, 
suppose that 
  $\vec{W}$ is such that $(M,\vec{u}) \sat \vec{W} = \vec{w}$ and for
some subset $\vec{Y}'$ of $\vec{Y}$, we have that
$(M,\vec{u}) \sat [\vec{Y}' \gets \vec{y}'', \vec{W} \gets
  \vec{w}]\neg \phi$, where $\vec{y}''$ is the restriction of
$\vec{y}'$ to the variables in $\vec{Y}'$. 
We can assume that $\vec{W}$ is a subset of
  $\vec{X}'$, since $\vec{X}'$ includes all the parents of the
  variables in $\phi$.  By causal independence, $(M,\vec{u}) \sat
  [\vec{Y}' \gets \vec{y}'](\vec{W} = \vec{w})$.  Thus, 
 $(M,\vec{u}) \sat [\vec{Y}' \gets \vec{y}'']\neg \phi$.  But this
  contradicts the minimality of $\vec{Y}$.  It follows that AC3 holds.

  We have shown that $\vec{Y} = \vec{y}$ is a cause of $\phi$ in
$(M,\vec{u})$.  Since $\vec{Y}$ includes a variable in $\vec{X}$, it
  follows that some conjunct of $\vec{X} = \vec{x}$ is part of a cause
  of $\phi$ in $(M,\vec{u})$, so SC2 holds, as desired. 
\eprf

%joe14*: This is misplaced.  MMTS don't define sufficient cause anywhere.
%\begin{proposition}
%  Theorem~\ref{theo:nodepend} does not hold for the MMTS's   definition.
%\end{proposition}

The notion of explanation builds on the notion of sufficient
causality, and is relative to a set of contexts.
\dfn\label{def:EX}[Explanation]
$\vec{X} = \vec{x}$ is 
%joe1
%an \emph{explanation} of $\varphi$ relative to a set of contexts $\K$
an \emph{explanation} of $\varphi$ relative to a set $\K$ of contexts 
in a causal model $M$ if the following conditions hold:  
\begin{description}
\item[{\rm EX1.}]  $\vec{X} = \vec{x}$ is a sufficient cause of $\varphi$ in all contexts in $\K$ satisfying 
%hana4
$(\vec{X}=\vec{x}) \wedge \varphi$. 
%$\vec{X}=\vec{x} \wedge \varphi$. 
More precisely,
\begin{itemize}
\item If $\vec{u} \in \K$ and $(M,\vec{u}) \models (\vec{X} = \vec{x})
  \wedge \varphi$, then there exists a conjunct $X=x$ of $\vec{X} =
  \vec{x}$ and a (possibly empty) conjunction $\vec{Y} = \vec{y}$ such
  that $X=x \wedge \vec{Y} = \vec{y}$ is an actual cause of $\varphi$
  in $(M,\vec{u})$. (This is SC2 applied to all contexts  
%joe1: this notation hasn't been defined
  %  $\vec{u} \in \K_{(\vec{X} = \vec{x}) \wedge \varphi}$).
    $\vec{u} \in \K$ where 
%hana4    
    $(\vec{X} = \vec{x}) \wedge \varphi$ holds.)
%    $\vec{X} = \vec{x}) \wedge \varphi$ holds.)
\item $(M,\vec{u}') \models [\vec{X} = \vec{x}]\varphi$  for all
  contexts $\vec{
    %joe6
%    u}' \in \K$. (This is SC3 restricted to the contexts in $\K$).
    u}' \in \K$. (This is SC3 restricted to the contexts in $\K$.)
\end{itemize}
%joe2
%\item[{\rm EX2.}] \label{ex2} $\vec{X}'$ is minimal; there is no
\item[{\rm EX2.}] $\vec{X}$ is minimal; there is no
  strict subset $\vec{X}'$ of $\vec{X}$ such that $\vec{X}' =
  \vec{x}'$ satisfies EX1,  
where $\vec{x}'$ is the restriction of $\vec{x}$ to the variables in $\vec{X}'$. (This is SC4).
%joe3
%\item[{\rm EX3.}] \label{ex3} $(M,u) \sat \vec{X} = \vec{x}) \wedge
\item[{\rm EX3.}] \label{ex3} $(M,u) \sat \vec{X} = \vec{x} \wedge
  \varphi$ for some $u \in \K$.
\end{description}
\edfn

%joe1*: let's cut this.  All explanations will be nontrivial in our
%setting, and it will just complicate things to carry this around
%\noindent The explanation is \emph{nontrivial} if it satisfies in addition
%\begin{description}
%\item[{\rm EX4.}] \label{ex4} $(M,\vec{u}) \models \neg{(\vec{X} =
%\vec{x})}$ for some $\vec{u}' \in \K_\varphi$. (The explanation is
%not already known given $\varphi$). 
%\end{description} 
%
%hana1*: non-triviality might be relevant when talking about
%explanations of rare outputs.  

%joe11
Note that this definition of explanation (which is taken from
\cite{Hal48}) is quite different from that in \cite{HP01b}.  
What is called EX2 in \cite{HP01b}  is actually an analogue of the
first part of EX1 here; although it is called ``sufficient
causality'', it is closer to the necessity condition.  But, like the
necessity clause used by MMTS,  it requires only that some
subset of $\vec{X} = \vec{x}$ be an actual cause of $\phi$ (without
allowing the subset to be extended by another conjunction $\vec{Y} =
\vec{y}$) (and uses a different definition of actual cause---that of
\cite{HP01a}).  The definition of \cite{HP01b} has no
analogue to the second part of EX1 here.

%joe14*: moved this discussion here
Of course, if the assumptions of Theorem~\ref{thm:nodepend} hold,
then we can drop the requirement that the first part of EX1 holds.
(Note that if the assumptions of Theorem~\ref{thm:nodepend} hold for
some context, they hold for all contexts; in particular, assumption
(d) holds, since $\vec{X}'$ includes all the parents of the variables
in $\phi$.)  Although the changes made by MMTS to Halpern's definition
seem minor, they are enough to prevent Theorem~\ref{thm:nodepend} from
holding.  We show this in Example~\ref{excat} in the next section,
after we have discussed the assumptions made by MMTS in more detail.

%joe1: slowing down; adding a sentence
%joe6
%The requirement that the first part of condition 1 holds in all
The requirement that the first part of condition EX1
%joe11
as given here
holds in all
contexts in $\K$ that satisfy 
$\vec{X} = \vec{x} \land \phi$ and that
the second part holds in all contexts in $\K$ is quite strong, and
often does not hold in practice.   We are often willing to accept
$\vec{X} = \vec{x}$ as an explanation if these requirements hold with
high probability.  
%joe1
%Given a causal model $M$ and a pair $(\K,Pr)$,
Given a set $\K$ of contexts in a causal model $M$, 
let $K_\psi$ consist of all contexts $\vec{u}$ in $\K$ such that
$(M,\vec{u}) \sat \psi$, and 
let $\K(\vec{X} =
\vec{x}, \varphi, \mbox{SC2})$ consist of all contexts $\vec{u} \in
\K$ that satisfy $\vec{X} = \vec{x}\wedge \varphi$ and the first
  condition in EX1 (i.e., the analogue of SC2).
%Formally,
%\[  \K(\vec{X} = \vec{x}, \varphi, \mbox{SC2}) = \{ \vec{u} \in \K_{(\vec{X} = \vec{x}) \wedge \varphi}: \exists \mbox{ a conjunct } X=x \mbox { of } vec{X} = \vec{x} \]
%\[ \;\;\;\;\;\;\; \mbox{and a (possibly empty) conjunction } \vec{Y}=\vec{y} : X=x \wedge \vec{Y}=\vec{y}  \mbox{ is a cause of } \varphi \mbox{ in } (M,\vec{u}) \}. \]
%joe1: not needed
%  Similarly, let $\K(\vec{X} = \vec{x}, \varphi, \mbox{SC3})$ consist of all
%contexts $\vec{u} \in \K$ that satisfy the second condition in EX1
%(i.e., the analogue of SC3); that is
%\[ \K(\vec{X} = \vec{x}, \varphi, \mbox{SC3}) = \{ \vec{u} \in \K:
%(M,\vec{u}) \models [\vec{X} \leftarrow \vec{x}]\varphi \}. \] 
\dfn\label{def:PEX}[Partial Explanation]
$\vec{X} = \vec{x}$ is 
a \emph{partial explanation} of $\varphi$ with goodness
%joe1
%$(\alpha,\beta)$ relative to $(\K,Pr)$ in a causal model $M$ if
$(\alpha,\beta)$ relative to $\K$ in a probabilistic causal model $(M,\Pr)$ if
%joe1*: I know that I have this in the book, but in retrospect, it
%seems too complicated and not quite right.
%$\vec{X}=\vec{x}$ is an explanation of $\varphi$ 
%relative to 
%\[ \K(\vec{X} = \vec{x}, \varphi, \mbox{SC3}) \setminus (\K_{\vec{X}
%= \vec{x} \wedge \varphi} \setminus \K(\vec{X} = \vec{x}, \varphi,
%\mbox{SC2})), \] 
%joe1: replace = by \le
%\[ \alpha = Pr(\K(\vec{X} = \vec{x}, \varphi, \mbox{SC2} | \K_{\vec{X}
\begin{description}

\item[{\rm EX1$'$.}]  
  $\alpha \le Pr(\K(\vec{X} = \vec{x}, \varphi, \mbox{SC2}) \mid
 \K_{\vec{X} = \vec{x} \land \phi})$ and
%joe1: again
%\[ \beta = Pr(\K(\vec{X} = \vec{x}, \varphi, \mbox{SC3})).\]
%hana7
 $\beta \le Pr(\K_{[\vec{X} = \vec{x}]\phi})$.
%$\beta \le Pr(\K_{[\vec{X} = \vec{x}]\phi]})$
\item[{\rm EX$2'$.}] $\vec{X}$ is minimal; there is no
  strict subset $\vec{X}'$ of $\vec{X}$ such that
  $\alpha \le Pr(\K(\vec{X}' = \vec{x}', \varphi, \mbox{SC2}) \mid
  \K_{\vec{X}' = \vec{x}' \land \phi})$
  and
  $\beta \le Pr(\K_{[\vec{X}' = \vec{x}']\varphi})$,
    %joe11: added 
  where $x'$ is the restriction of $\vec{x}$ to the variables in $X$.
\item[{\rm EX$3'$.}]   $(M,u) \sat \vec{X} = \vec{x} \wedge
  \varphi$ for some $u \in \K$.
\end{description}  
\edfn

%hana12 added
%joe17
%There is no natural counterpart to Theorem~\ref{thm:nodepend} for
%partial explanations, as the conjuncts in
%EX1$'$ can be satisfied by different contexts in the set of contexts $\K$.
%The theorem would hold if both conjuncts are satisfied
%by the same subset of $\K$ (and hence also $\alpha=\beta$). 
Theorem~\ref{thm:nodepend} has no obvious counterpart for
partial explanations.  The problem is that the conjuncts in 
EX1$'$ can be satisfied by different contexts in the set $\K$, while still
satisfying the probabilistic constraint.
(The theorem would hold with $\alpha = \beta$ if both conjuncts were
satisfied by the same subset of $\K$.)

%joe6
%\section{Explanations of Image Classifiers in the Framework of Actual
%  Causality}\label{sec:imagexp}
\section{Using Causality to Explain Image Classification}\label{sec:imagexp} 

%joe1
%We view the classifier (a neural network) as an unknown causal model
%joe14: rewrote to make it clear that we're following MMTS here
%We view an image classifier (a neural network) as a probabilistic causal model.
%We take the endogenous variables to be the set $\vec{V}$ of pixels
Following MMTS, we view an image classifier (a neural network) as a
probabilistic causal model.  MMTS make a number of
%joe14
additional
assumptions in
their analysis.
%joe14*
%which we make as well, unless we explicitly say otherwise.
Specifically, MMTS take
the endogenous variables to be the set $\vec{V}$ of pixels
that the image classifier gets as input, together with an output
%joe14
%variable $O$.  The variable $V_i \in \vec{V}$ describes the color and
variable that we call $O$.  The variable $V_i \in \vec{V}$ describes
the color and 
%joe14
%intensity of pixel $i$; it's value is determined by the exogenous variables.  
intensity of pixel $i$; its value is determined by the exogenous variables.  
The equation for $O$ determines the output of the
neural network as a function of the pixel values.
%joe14
Thus, the causal network has depth 2, with the exogenous variables
determining the feature 
%hana10
variables,
%values, 
and the feature variables determining
the output variable.  There are no dependencies between the feature
variables.  Moreover, for each setting $\vec{v}$ of the feature
variables, there is a setting of the exogenous variables such that $\vec{V}
= \vec{v}$.
%joe14
%That is, the feature vectors are $\vec{V}$-independent,
That is, the variables in $\vec{V}$ are causally independent and
determined by the context,
in the sense of Theorem~\ref{thm:nodepend}.  Moreover, all the parents
of the output variable $O$ are contained in $\vec{V}$.  So for any
explanation of the output involving the pixels, conditions (a), (b),
(c), and (e) of Theorem~\ref{thm:nodepend} hold if we take $\phi$ to be
some setting of $O$.
While a neural
%joe14
%network often outputs several labels, we assume for ease of exposition
network often outputs several labels, MMTS assume 
that the output is unique (and a deterministic function of the pixel
%joe6: this is backwards!
%values).  Finally, we assume that the context $\vec{u}$ determines the
%pixel values, so that for each image $I$ there is a unique context
%$\vec{u}_I$.  The probability on contexts corresponds to the
values).
%joe14: MMTS don't assume this; let's get rid of it.
%Finally, we assume that for each image $I$ there is a unique context
%$\vec{u}_I$ that gives rise to image $I$.
%The probability on contexts then
Given these assumptions, the probability on contexts directly
corresponds to the 
probability on seeing various images (which the neural network
presumably learns during training).
%joe14*: I don't think we need this
%We assume that the classifier is
%nontrivial in the sense that there are at least two labels $o_1$ and
%$o_2$ such that some images get label $o_1$ and some images g et label
%$o_2$ (i.e., it is not the case that all images get the same label).

%joe4*: MMTS actually assumed both of these conditions (they must have
%read my mind in advance :-)).  So I rewrote this to say that
%We now want to provide a natural sufficient condition under which, for
%joe6*: rewrote
%Although they said that they were using Halpern's definition, the
%definition of explanation that MMTS actually made two simplifying
Although they claimed to be using using Halpern's definition, the
definition of (partial) explanation given in MMTS differs from that of
Halpern in three respects.  The first is that, rather than requiring that a
conjunct of the explanation can be extended to an actual cause (i.e.,
the first part of EX1), they require that in each context, some subset
of the explanation be an actual cause.  Second, they do not use 
Halpern's definition of actual cause; in particular, in their analogue of AC2,
they do not require that $\vec{W} = \vec{w}$ be true in the
context being considered.  This appears to be an oversight on their
part, and is mitigated by the fact that they focus on but-for
causality (for which $\vec{W} = \emptyset$, as we observed, so that
the requirement that $\vec{W} = \vec{w}$ in the context being
considered has no bite).  
%hana4* is it the set of images or the set of exogenous variables? if
%it is the set of images (which I think it should be), it is not
%defined
%joe7: \K has to be a subset of \U, so it's a set of contexts.  Since
%we (and MMTS, I think) identify contexts with images, this means that
%they're considering all images.  
%joe8
%Finally, they take $\K = \U$.
Finally, they take $\K = \R(\U)$.
%joe7: added next sentence
Since we have identified contexts and images, this amounts to
considering all images possible.
As we shall
see in Section~\ref{sec:beyond}, there are some benefits in allowing
%joe8
%the greater generality of having $\K$ be an arbitrary subset of $\U$.
the greater generality of having $\K$ be an arbitrary subset of $\R(\U)$.  
%assumptions relative to Halpern's definition of what it means for
%$\vec{X} = \vec{x}$ to be an explanation of $O=o$
%(i.e., the output label $o$) with goodness $(\alpha,\beta)$ relative
%%joe6
%%to the set $\K$ of all contexts.  One
%condition is obvious: we want setting $\vec{X} = \vec{x}$ to be a
%sufficient condition for getting output label $o$ with probability at
%least $\beta$; that is, $\beta \le Pr(\K_{[\vec{X} =
%    \vec{x}](O=o)]})$. The condition that we need to get the first
%  part of EX1 (i.e., the analogue of SC2) to hold is also quite
%  intuitive.
%joe3*: new condition
%  Let $\vec{Y} = \vec{V} - \vec{X}$.  If we restrict to
%  contexts where $\vec{X} = \vec{x}$ and $O=o$, we want it to be the
%  case that, with probability at least $\alpha$, changing the values
  %  of the pixels in $Y$ does not change the output.
%It simply says that if we change enough of the pixel values in
%$\vec{X}$, we will get a different output.  
%joe4
%joe6*
%This essentially amounts to but-for causality.  MMTS explicitly noted
%that they were simplifying things by assuming but-for causality, but
%they did not note the first simplification (in part, because they
%never explicitly gave Halpern's definition).
%joe6*
%We start by showing that these conditions really are sufficient to
%give Halpern's notion.
We now show that the conditions considered by MMTS suffice to give
Halpern's notion.

\begin{theorem}\label{thm:partialexplanation}
    For the causal model $(M,\Pr)$ corresponding to the image
  classifier, $\vec{X} = \vec{x}$ is a partial explanation of $O=o$
%joe14*
%  with goodness $(\alpha,\beta)$, where $\beta > 0$ if the following
  with goodness $(\alpha,\beta)$, where $\alpha, \beta > 0$, if the following
  conditions hold:
  \begin{itemize}
  \item $\beta \le Pr(\K_{[\vec{X} = \vec{x}](O=o)]})$;
%joe14 switched second and third coditions, to make it match the proof.
  \item there is no strict subset $\vec{X}'$ of $\vec{X}$ such that 
%joe3
      %      $\beta \le Pr(\K_{[\vec{X'} = \vec{x'}](O=o)]})$.
      $\beta \le Pr(\K_{[\vec{X}' = \vec{x}'](O=o)]})$,
  %joe11: added 
      where $x'$ is the restriction of $\vec{x}$ to the variables in $X$;
        \item $\alpha \le \Pr(\{\vec{u}: \exists \vec{x}'( (M,\vec{u}) \sat
    [\vec{X} = \vec{x}'](O \ne o))\} \mid \K_{\vec{X} = \vec{x} \land
          O=o})$.
  \end{itemize}
\end{theorem}
  
\prf  The first condition in the theorem clearly guarantees
  that the second part of EX1$'$ holds; the second condition
  guarantees that EX2$'$ holds.  The fact that $\beta > 0$ means that for
  some $\vec{u}$, we must have $(M,\vec{u}) \sat [\vec{X} \gets
    %hana4* I think it is not \phi but O=o?
    %joe7: yes; that's right.
%joe3
%    \vec{x}]]\phi$.  Let $\vec{u}'$ be the context that gives the
%joe14: no longer unique
%    \vec{x}](O=o)$.  Let $\vec{u}'$ be the context that gives the
    \vec{x}](O=o)$.  Let $\vec{u}'$ be a context that gives the
    %\vec{x}]\phi$.  Let $\vec{u}'$ be the context that gives the
    pixels in $\vec{X}$ the value $\vec{x}$ and agrees with $\vec{u}$
    on the pixels in $\vec{Y}$.  Clearly $(M,\vec{u}') \sat \vec{X} =
    \vec{x} \land O=o$, so EX3$'$ holds.

    It remains to show that the first part of EX1$'$ holds.
Suppose that  $\vec{u}$ is a context for which there exists a setting
$\vec{x}'$ of the pixels in $\vec{X}$ such that $(M,\vec{u}) \sat
[\vec{X} = \vec{x}'](O \ne o)$.
%joe14
(Since $\alpha > 0$, there must exist such a context.)
Let $\vec{X}'$ be a minimal subset
of $\vec{X}$ for which there exists a setting $\vec{x}''$ such that 
$(M,\vec{u}) \sat [\vec{X}' = \vec{x}''](O \ne o)$.  It is easy to
see that $\vec{X}' = \vec{x}^*$ is a cause of $O=o$, where $\vec{x}^*$
is the restriction of $\vec{x}$ to $\vec{X}'$.  Thus, the first part
of EX1$'$ holds for $\vec{u}$.  The desired result follows.
\eprf
    \commentout{
    Let $\U' = \{\vec{u}: \forall \vec{y} (M,\vec{u}) \sat
    [\vec{Y} = \vec{y}](O = o)\}$. 
    Given a context $\vec{u} \in \K_{\vec{X} = \vec{x}
      \land O=o}$, suppose that there is a minimal set $\vec{Z}$ of
    pixels that includes at least one pixel in $\vec{X}$ such that
    changing the values of the pixels in $\vec{Z}$ from their value in
    $\vec{u}$ results in a label other than $o$.  If $(M,\vec{u}) \sat
    \vec{Z} = \vec{z}$, then it is easy to see that $\vec{Z} =
    \vec{z}$ is a cause of $O=o$ in $(M,\vec{u})$; moreover, we can 
    write $\vec{Z} = \vec{z}$ as $X = x \land \vec{Z}' = \vec{z}'$ for
    some pixel $X \in \vec{X}$, where $\vec{Z}' = \vec{Z} - \{X\}$.
    Thus, the first part of EX1 holds for $\vec{u}$.  On the other
    hand, if we can't do this, then $\vec{Z} \subseteq \vec{Y}$, so
    there is a setting $\vec{y}$ of the variables in $\vec{Y}$ such
    that $(M,\vec{u}) \sat [\vec{Y} = \vec{y}](O \ne o)$; that is,
    $\vec{u} \notin \U'$.  By assumption $\Pr(\U')\mid \K_{\vec{X} =
      \vec{x} \land \phi} \ge \alpha$.  it follows that the set of
    contexts $\vec{u}$ for which the first part of EX1 holds has
    probability at least $\alpha$, conditional on $\K_{\vec{X} =
      \vec{x} \land \phi}$, as desired.}

%and a context $\vec{u}$ gives these pixels their current values,
%representing color and intensity.
%and a context $\vec{u}$ gives these pixels their current values,
%representing color and intensity.
%
%An input image $I$ is defined by 
%$\vec{V}$ and $\vec{u}$. (that is, an input image $I$ is defined by
%the context $\vec{u}$).
%While a classifier typically outputs several labels, for simplicity
%we (can be the one with the highest probability) $O$, whose value is the
%label assigned to the input. 
%Let the current label be $o$ and the formula $\varphi = (O=o)$, that
%is $(M,\vec{u}) \models O=o$.  

%A part of an input image is a subset $\vec{X}$ of $\vec{V}$ with the
%values $\vec{x}$ defined by $\vec{u}$.  A partially obscured image 
%$I'(\vec{X})$ 
%is defined as an image that agrees with $I$ on the values of
%$\vec{X}$, and all its other pixels having a special value $d$
%(default, the color of 
%the occlusion). Let $\vec{u}_d$ denote a context that assigns all
%pixels of the input the default value $d$. 

%joe4*: added discussion and rewrote
%joe3*: new material
%While the second condition in Theorem~\ref{thm:partialexplanation}
%seems quite natural, it is instructive to see examples where it does
%not hold (and we do not get partial explanations).  
%joe9
%How reasonable are these assumptions?

How reasonable are the assumptions made by MMTS?
%hana6*
%joe9*: The issue is *not* but-for causality, but the first assumption
%made by MMTS.  The discussion in the example is also incorrect in
%other ways.
%We claim that considering just but-for causality is not reasonable
%even if we assume that there are no causal connections between the
%setting of different variables (or various pixels). The following
%example illustrates this claim.
The following example shows that their assumption that, in each
context, some subset of the explanation be an actual cause (as opposed
to some conjunct of the explanation being extendable to an actual
cause, as required to EX1) leads to
arguably unreasonable explanations.

\begin{example}\label{excat}
%joe9
%Consider the following voting setting. Suppose there are three voters
Consider the following voting scenario. There are three voters,
  $A$, $B$, and $C$, 
%joe9
%who can vote for the candidate or abstain,  and just one vote is
who can vote for the candidate or abstain; just one vote is
needed for the candidate to win the election.
%joe14
The voters make their decisions independently.
%joe9: actually, we don't need this.
%In the actual context  $\vec{u}$,  all three voters vote for the candidate.
By Definition~\ref{def:EX},
%joe9
 % $A$ is an explanation of the outcome
  $A=1$ (the fact that $A$ voted for the candidate) is an explanation
 of the outcome  
%joe9
 % (as well as $B$ and $C$, separately). Indeed, it is easy to see that
  (as well as $B=1$ and $C=1$, separately). By assumption,
 %joe9
 $A=1$ is sufficient for the candidate to win in all contexts.
 Now consider any context $\vec{u}$ where the candidate wins and $A=1$.
 It is easy to see that the conjunction of voters for the candidate in
 $\vec{u}$ is a cause of the candidate winning.  So, for example, if
 $A$ and $C$ voted for the candidate in context $\vec{u}$, but $B$ did
 not, then $A=1 \land C=1$ is a (but-for) cause of the candidate
 winning; if both votes change, the candidate loses.
 %joe1
 % $A \wedge B \wedge C$
%  present in the actual cause. Moreover, $A$ is sufficient for $G$'s

On the other hand, $A=1$ is \emph{not} an explanation of the
candidate winning according to the MMTS definition.  To see why, note
that in the context $\vec{u}$ above, no subset of $A=1$ is an actual
cause of the candidate winning.  Rather, according to the MMTS
%joe10
%definition,   $A=1 \wedge B=1 \wedge C=1$ is the only explanation
definition,   $A=1 \wedge B=1 \wedge C=1$ is the only explanation
(since there are contexts---namely, ones where all three voters voted
for the candidate)---where $A=1 \wedge B=1 \wedge C=1$ is the only
actual cause).
% present in the actual cause. Moreover, $A$ is sufficient for $G$'s
% win. This also seems to match   our intuition.
% However, $A$ is not an explanation according to MMTS -- only the
% conjunction of all  three votes is.
This does not match our intuition.  (Of course, we can easily convert
this to a story about image classification, where the output is 1 if
any of the pixels $A$, $B$, and $C$ fire.)

%joe14*
As we observed, the assumptions of MMTS imply that conditions (a), (b),
(c), and (e) of Theorem~\ref{thm:nodepend} hold, if we take $\vec{X}'$ to
be the pixels, $\vec{X}$ to be some subset of pixels, and $\phi$ to be some
setting of the output variable $O$.  They also hold in this example,
taking $\vec{X}'$ to be the set of voters, $\vec{X}$ to be any subset
of voters, and $\phi$ to be the outcome of the election.  Moreover,
condition (d) holds.  So in this example, we do not need to check the
first part of EX1 to show that $A=a$ is an explanation of the
candidate winning, given that the second part of EX1 and EX2 hold.
But this is not the case for the MMTS definition.  Although $A=1$ is a
sufficient cause of the candidate winning and is certainly minimal, as
we observed, it is not an explanation of the outcome according to the
MMTS definition.  This is because MMTS require a subset of the
conjuncts in the explanation to be an actual cause, which is not the
case for SC2.

%joe14
%A similar situation occurs in explaining image classification. As
% observed in~\cite{SLKTF21,CKK23}, images usually 
For another, perhaps more realistic, example of this phenomenon in
image classification, as observed by Shitole et al.
%joe18
%\citeyear{SLKTF21} and Chockler et al. \citeyear{CKK23}, images usually 
\nciteyear{SLKTF21} and Chockler et al. \nciteyear{CKK23}, images usually 
have more than one explanation. Assume, for example, that the input
 image $I$ is an image of a cat, labeled as a cat 
 by the image classifier. An explainability tool might find several
 explanations for this label, such as the cat's ears and nose, 
 the cat's tail and hind paws, or the cat's front paws and fur. All
 those are perfectly acceptable as explanations of 
%joe9
% why this image was labeled a ``cat'', but only their union is an
%joe14
% why this image was labeled a ``cat'', but only their conjunction is an
 why this image was labeled a cat, but only their conjunction is an
 explanation according to MMTS.
%joe11*: Hana, what do LIME and SHAP call an explanation in this case?
%DSo they agree with my definition?
%hana7* LIME and SHAP only find one explanation. 
 %So in this case, they will find, say, cat's ears and nose.
 %joe12*: So is there explanation more like MMTS or like mine?  That
 %is, it is a conjunction (of what we would call the possible explanations).
 %Is it sometimes just one conjunct?  If so, what does it depend on?
 %It would be good to say something here, if we can say something useful.
 %hana8*
 In contrast, most of the existing explainability tools for image
%joe13
 % classifiers output one explanation.
  classifiers output one explanation (i.e., a single conjunct).   
%joe13
%  These explanations match Definition~\ref{def:EX} for the set of
Thus, these explanations match Definition~\ref{def:EX} (rather than
that of MMTS) for the set of
  contexts that includes the original image and all 
 partial coverings of this image.
\end{example}
 
 %hana6*
%Considering just but-for
%causality is reasonable
%joe9: their definition does not assume but-for causality.
%joe13
%MMTS' restriction to but-for causality
MMTS's restriction to but-for causality 
%joe9: reinstated
is reasonable if we assume that there are no causal
connections between the setting of various pixels.   However,
there are many
examples in the literature showing that but-for causality does not
suffice if we have a richer causal structure.  Consider the following
well-known example due to Hall \nciteyear{Hall98}:

\begin{example}\label{ex1}
Suzy and Billy both pick up rocks and throw them at a bottle. Suzy’s
rock gets there first, shattering the bottle. Since both throws are
%joe6
%perfectly accurate, Billy’s would have shattered the bottle had it not
perfectly accurate, Billy’s rock would have shattered the bottle had it not
been preempted by Suzy’s throw.  The standard causal model for this
%hana4 fixing the reference
%joe6
%story (see \cite{HP01a} has endogenous binary variables $ST$ (Suzy
story (see \cite{HP01b}) has endogenous binary variables $ST$ (Suzy
%story (see \cite{HP01a}) has endogenous binary variables $ST$ (Suzy
throws), $BT$ (Billy throws), SH (Suzy's rock hits the bottle), $BH$
%joe6
%(Billy's rock hits the bottle), and $BS$ (the bottle shatters.  The
(Billy's rock hits the bottle), and $BS$ (the bottle shatters).  The
values of $ST$ and $BT$ are determined by the exogenous variable(s).
The remaining equations are $SH = ST$ (Suzy's rock hits the bottle if
Suzy throws), $BH = BT \land \neg BH$ (Billy's rock hits the bottle if
Billy throws and Suzy's rock does not hit the bottle---this is how we
%joe6
%capture that Suzy's rock hits the bottle first), and $BS = SH \lor BH$
capture the fact that Suzy's rock hits the bottle first), and $BS = SH \lor BH$
%joe6
%(the bottle shatters if it is hit by either rock.
(the bottle shatters if it is hit by either rock).  

%joe6
%Suppose that $\L$ consists of four contexts, corresponding to the four
Suppose that $\K$ consists of four contexts, corresponding to the four
possible combinations of Suzy throwing/not throwing and Billy
throwing/not throwing, and $\Pr$ is such that all context have
positive probability.    In this model, Suzy's throw (i.e., $ST = 1$)
is an explanation for the bottle shattering (with goodness (1,1)).
Clearly the second part of 
EX1 holds; if Suzy throws, the bottle shatters, independent of what
Billy does.  The first part of EX1 holds because $ST=1$ is a cause of
%joe6
%the bottle shattering; if we hold $BT$ fixed at 0 (it's actual value),
the bottle shattering; if we hold $BT$ fixed at 0 (its actual value),
then switching $ST$ from 1 to 0 results in the bottle not shattering.
But ST=1 is not a but-for cause of $BS=1$.  Switching $ST$ to 0 still
results in $BS=1$, because if Suzy doesn't throw, Billy's rock will
hit the bottle.

%joe6
%Now we can convert this story to a story more appropriate for image
We can easily convert this story to a story more appropriate for image
classification, where we have an isomorphic model.  Suppose that we
have two coupled pixels.  $ST$ and $BT$ correspond to whether we turn
%joe6
%on the power to the pixels.  However, if we set the first one (which
%corresponds to $SH$) is on, the second (which corresponds to $BH$) is
on the power to the pixels.  However, if we turn the first one (which
corresponds to $SH$) on, the second (which corresponds to $BH$) is
turned off, 
even if the power is on. We classify the image as ``active'' if
either pixel is on.  Since this model is isomorphic to the Suzy-Billy
story, turning on the first pixel is an explanation for the
classification, but again, the second condition of
Theorem~\ref{thm:partialexplanation} does not hold.
 %joe11*: Hana, again, how would LIME and SHAP deal with this example?
 %hana7* LIME, SHAP, and ReX all replace parts of an input image with
 %a background colour and query the classifier. 
%all assume independence between pixels.
%joe12*: This doesn't really answer my question though.  What would
%LIM/SHAP/REX call the explanation in this case.  Would they call
%turning on the first pixel an explanation?  
%hana8
%joe13
%Indeed, he explainability tools for image classifiers would output
%hana9
Given a context (an input image) in which the first pixel is on, the
%The 
explainability tools for image classifiers would output
the first pixel being on as an explanation 
for the classification. This is because they do not take the
dependencies between pixels into account.
%hana9
%joe14
%Note that the second pixel is not a part of an explanation, as it is
These systems would not call the second pixel part of an
explanation, as it is 
off in the 
input image.
%joe13*: Not taking dependecies into account doesn't feel like the
%right explanaation for the explanation.  I guess the question here is
%why they would *not* output the second pixel as an explanation if
%they don't take dependencies into acount.
 \end{example}

The following example shows that even without assuming causal structure
between the pixels, the second condition in
Theorem~\ref{thm:partialexplanation}  may not hold.

%joe3*: new example
\begin{example}\label{ex2}  Suppose that pixels have values in $\{0,1\}$.  Let an
%joe6
%  image be an $(2n+1)$-tuple of pixels.  Suppose that an image is
  image be a $(2n+1)$-tuple of pixels.  Suppose that an image is
%joe6
%  labeled 0 if the first pixel is a 0 and there are an even number of
%  0s in the remaining pixels, or the first pixel is a 1, and there are
%  a positive even number of 0s in the remaining pixels.  Suppose the
  labeled 0 if the first pixel is a 0 and the number of 0s in the
  remaining $2n$ pixels is even   (possibly 0), or the first pixel
  is a 1, and the number of 0s in the remaining pixels is even and positive.
  Suppose that the
  probability distribution is such that the set of images where there
%joe6
%  are an even number of 0s in the final $2n$ pixels has probability .9,
%  uniformly distributed   among all tuples where there are an even
%  number of 1's in the final $2n$ pixels.
  is an even number of 0s in the final $2n$ pixels has probability $.9$,
  %joe6
  Moreover, suppose that $X_1 = 0$ with probability $1/2$, where $X_1$
  denotes the first pixel.  
  Now the question is whether
  $X_1 = 0$ is a partial explanation of $O=0$ with goodness
  %joe6*
%  $(\alpha,.9)$ for some $\alpha > 0$, where $X_1$ is the first pixel.
  $(\alpha,.9)$ for some $\alpha > 1/2^{2n-1}$.
  Clearly the probability that $O=0$ conditional on $X_1 = 0$
%joe6
%  is .9, while unconditionally, the probability that $O=0$ is less
  is .9, while conditional on $X_1 = 1$ and unconditionally, the
  probability that $O=0$ is less 
  than $.9.$  Thus, the second part of EX$1'$ and EX2$'$ both hold.  
  Now consider the first part of EX1$'$.  Suppose that $(M,\vec{u})
  \sat X_1 = 0 \land O=0$, so in $\vec{u}$, an even number of the
  last $2n$ pixels are $0$.  Suppose in fact that a positive number of
  the last $2n$ pixels are $0$.  Then we claim that there is no
  $\vec{Y}$ and $\vec{y}$ such that $X_1 = 0 \land \vec{Y} = \vec{y}$
  is a cause of $O=0$ in $(M,\vec{u})$.  Clearly, $X_1 = 0$ is not a
  cause (since setting it to $1$ does not affect the labeling).
  Moreover, if $\vec{Y}$ is nonempty, let $Y \in \vec{Y}$ and let $y$
  be the value of $Y$ in $\vec{Y}$.  Then it is easy to see that
  $Y = y$ is a cause of $O=0$ (flipping the value of $Y$ results in
  changing the value of $O$ to $1$), so $X_1 = 0 \land \vec{Y} =
  \vec{y}$ is not a cause of $O=o$ (AC3 is violated).  Thus, $X_1 = 0$
  is a cause of $O=0$ only in the context $\vec{u}$ where
  %joe6
% all the   last $2n$ pixels are 1.  This context has probability $1/n+1$
  $X_1 = 0$ and 
 all the   last $2n$ pixels are $1$.  Using Pascal's triangle, it is
 easy to show that this context has probability $1/2^{2n-1}$
  conditional on $X_1 = 0 \land O=0$.  Note that this is also the only
  context where changing the value of $X_1$ affects the value of $O$.
  \end{example}

%joe4: added discussion
Example~\ref{ex2} is admittedly somewhat contrived; it does not seem
that there are that many interesting examples of problems that arise
if there is really no causal structure among the pixels.  But, as
Example~\ref{ex1} suggests, there may well be some causal connection
between pixels in an image.  Unfortunately, none of the current
approaches to explanation seems to deal with this causal structure.
We propose this as an exciting
area for future research; good explanations will need to take this
causal structure into account.

%hana4* Actually, none of the existing explanations of image
%classifiers is a but-for explanation, because for each image, there
%are usually 
%several explanations. So covering one of them does not change the classification. For example, one explanation of a cat is its head. However,
%covering the cat's head in the image does not change the classification, because a cat can also be identified by, say, its paws, fur, and tail.
%joe7: note that it's but-for causality, not but-for explanation.  The
%claim is also on a per-context basis. So in a particular context, it
%might already be the case that there is nothing that could be the
%paws, fur, or tail.  And if we don't know that, then the part that
%could be tail, for example, would be part of the cause, and we could
%change that too.  Bottom line: I still think what we said is right,
%and we lose something by insisting on but-for causality.

%joe4*: this is subsumed by MMTS; replaced it by the discussion above
\commentout{

%hana1
%joe2
%\paragraph{Small explanations}
\paragraph{Basic explanations}
%\paragraph{Minimal explanations}
To what extent do the various approaches that claim to provide 
%hana1
basic
%minimal
explanations really provide an explanation in a more formal sense?
Consider ReX.
Given an image $I$ and a set $\vec{X}$ of pixels, let $\vec{x}_{I}$
be the values of the pixels in $\vec{X}$ in the image $I$.
Let $I_{\vec{X}}$ be the
\emph{partially obscured image}  that gives the pixels in $\vec{X}$
the value $\vec{x}_I$ and all the pixels in $\vec{V} - \vec{X}$ the special
default value $d$.  ReX takes $\vec{X}=\vec{x}_I$ to be an
explanation of the labeling of an image $I$ if $I_{\vec{X}}$ has the
same label as $I$ and for no strict subset $\vec{X}'$ of $\vec{X}$ is it the
case that $I_{\vec{X}'}$ has the same label as $I$.
This would in fact be a partial explanation of goodness
$(\alpha,\beta)$ if the fact that $I_{\vec{X}}$ has label $o$ means
that with probability at least $\beta$, an image for which $\vec{X} =
\vec{x}_I$ will get label $o$ and, moreover, for images
%hana1* I think this is crucial
in the dataset
that have
$\vec{X} = \vec{x}_I$ and get label $o$,  with probability at least
$\beta$, changing any collection of pixels not in $\vec{X}$ will not
affect the label.  The implicit hope in ReX is that giving a
default value to all the pixels not in $\vec{X}$ results in an image
that is a good proxy for all images that have $\vec{X} = \vec{x}_I$.

%joe3
%[[HANA, CAN WE SAY SOMETHING SIMILAR ABOUT SHAP AND LIME.]]

%hana1
%joe3: Hana, we haven't talked at all about how DeepCover produces a
%ranking, and how this leads to an explanation for DeepCover.  If
%we're going to say that LIME and SHAP differ in how they produce the
%ranking, we need to say this.  
LIME and SHAP differ from ReX in the way they produce the
ranking from which 
an explanation is extracted.
%joe3: cut; this isn't relevant to our discussion
%and also in the additional information
%they provide to the 
%user: LIME provides the simple neural network it used to produce the
%ranking, and 
%SHAP provides the Shapley values of the pixels.
However, with respect to the visual 
explanations, our description of the ReX output above holds for
LIME and SHAP  as well.
%joe3*: Hana, we still need to say more.  I'm confused about SHAP, for
%example,  You said earlier "the subset of pixels with the hightest Shapley
%value is then presented as the explanation."  Nowhere do you talk
%about setting values to a default with SHAP.  I don't see where that
%fits in.  This needs to be clarified.  Likewise for LIME.  
%In addition, it would be very useful to comment on the extent to
%which DeepCover, Lime, and SHAP satisfy the new condition in the theorem.
}
%joe4: \end{commentout}

%joe4*: tried to make this a better story.
\section{Beyond Basic Explanations}\label{sec:beyond}

%hana3 I don't think it is necessary
\commentout{
In this section, we discuss 
%In this section, we consider a variant of the basic notion of
%explanation considered above, namely, \emph{robust explanation}, and
two issues that arise when dealing with explanations: explanations of
absence and dealing with rare events.  While explanations of absence
are of great practical interest, 
} %end of commented out text

%hana3 
\commentout{
\paragraph{Robust explanation:}
We say that $\vec{X} = \vec{x}$ is a \emph{robust explanation}
of $O=o$ with goodness $(\alpha,\beta)$ in $(M,\Pr)$ if, for all
subsets $\vec{Y}$ of $\vec{V} - \vec{X}$ and all setting $\vec{y}'$ of
$\vec{Y}$, $\vec{X} = \vec{x}$ is an explanation
of $O=o$ with goodness $(\alpha,\beta)$ in $(M,\Pr \mid \vec{Y} =
\vec{y}')$.  Note that by conditioning $\Pr$ on $\vec{Y}' = \vec{y}'$, we
are effectively considering what happens when we set $\vec{Y}'$ to
$\vec{y}'$.  So this is saying that, no matter how we fix parts of the
image outside of $\vec{X}$, $\vec{X} = \vec{x}$ provides an
explanation of goodness $(\alpha,\beta)$ for $O=o$.
%hana1*: actually, this is not entirely accurate. Anchors superimposes another image
%from the dataset over the rest of the pixels. So the values of other pixels are
%not completely arbitrary. 
%joe3: is this right
%[[HANA, NOW IT WOULD BE GOOD TO SAY SOMETHING ABOUT THE EXTENT TO
%    WHICH ANCHORS PROVIDES A ROBUST EXPLANATION IN THIS SENSE.]]
Anchors views $\vec{X} = \vec{x}$ as an explanation if we get the same
label no matter how we replace the remaining pixels by another image
from the dataset, so it is trying to provide robust explanations.
%hana2
Trying all possible replacements of the pixels in $V - \vec{X}$ with pixels from another image
in the dataset is infeasible for large datasets. Instead, Anchors
\emph{samples} the dataset by 
%joe5: I don't understand this.  Do they have a bound and try that
%many?  Where did the bound come from?
trying a specific number of images. 
} %end of commented out text

%joe4*:
%how does Anchors deal with explosion (i.e., the need to consider consider all subsets $\vec{Y}$ of
%    $V - \vec{X}$))?

%hana1
%joe3: moved it earlier
%The resulting explanation provided by Anchors is a part of the image
%that is sufficient to guarantee the same classification regardless of
%the values of the rest of the pixels.    

%joe1*: cut this definition.  It doesn't seem necessary nor does MEX2
%make sense.
\commentout{
\dfn\label{def:minimex}
Given an image classifier $M$ and an input image $I$ defined by the context $\vec{u}$, a subset of variables $\vec{X}$ of $M$ with the values $\vec{x}$
in $(M,\vec{u})$is a \emph{minimal explanation} of the classification $O=o$ of $I$ if the following hold:
\begin{description}
\item[{\rm MEX1.}] $(M,\vec{u}) \models (\vec{X}=\vec{x}) \wedge (O=o)$.
%joe1: Did you mean (M,\vec{u}_d) |= [\vec{X} = \vec{x}](O=0)?
\item[{\rm MEX2.}] $(M,\vec{u}_d)[\vec{X}=\vec{x}] \models O=o$.
\item[{\rm MEX3.}] $\vec{X}=\vec{x}$ is minimal. That is, there is no $\vec{X}' \subset \vec{X}$ with the values $\vec{x}'$ that satisfies MEX2.
\end{description}
\edfn

%joe1*: This lemma doesn't make sense.  Definition 4 has as
%probability; there is no probability in Definition 5.  I don't think
\begin{lemma}
Definition~\ref{def:minimex} is equivalent to Definition~\ref{def:PEX} for image classifiers.
\end{lemma}

Let us now formalize the definition of a robust explanation. We define the set of contexts $\K$ as the set of all possible images in the dataset. 
\dfn\label{def:robimex}
Given an image classifier $M$ and an input image $I$ defined by the context $\vec{u}$, a subset of variables $\vec{X}$ of $M$ with the values $\vec{x}$
in $(M,\vec{u})$is a \emph{minimal robust explanation} of the classification $O=o$ of $I$ with respect to the set of contexts $\K$ if the following hold:
\begin{description}
\item[{\rm REX1.}] $(M,\vec{u}) \models (\vec{X}=\vec{x}) \wedge (O=o)$.
\item[{\rm REX2.}] $(M,\vec{u}')[\vec{X}=\vec{x}] \models O=o$ for all $\vec{u}' \in \K$.
\item[{\rm REX3.}] $\vec{X}=\vec{x}$ is minimal. That is, there is no $\vec{X}' \subset \vec{X}$ with the values $\vec{x}'$ that satisfies REX2.
\end{description}
\edfn

%joe1*: this also doesn't seem true
\begin{lemma}
Definition~\ref{def:robimex} is equivalent to Definition~\ref{def:EX} for image classifiers.
\end{lemma}
}
%joe1: \end{commentout}

%joe2
%\section{Explanations of Absence}
%hana3 cut the paragraph for now, as there is no other material in this section
%\paragraph{Explanations of Absence and Rare Events}

%joe1
%So far all explanation tools examined classifiers that output only
So far we have considered classifiers that output only
positive labels, that is, labels that describe the image. However,
there are classifiers 
that output negative answers. Consider, for example,
the use of AI in healthcare, where image classifiers are used as a
part of the diagnostic procedure for MRI images, X-rays,
%joe1
%mammograms  etc.~\cite{APR2019,PCR20}. In the use of image classifiers
and mammograms \cite{APR2019,PCR20}. In the use of image classifiers
in brain tumor detection based on an MRI, the classifier outputs 
either ``tumor'' or ``no tumor''.
%joe1
%While we understand how would an explanation for a tumor look like,
%based on the definitions in 
%Section~\ref{sec:imagexp}, it is less clear what would be an
%explanation for the ``no tumor'' label.
While the discussion above shows what would it would take for $\vec{X}
= \vec{x}$ to be an explanation of
``tumor'', what would count for $\vec{X} = \vec{x}$ to be an explanation of ``no tumor''?  We
claim that actually the same ideas apply to explanations of ``no
tumor'' as to tumor.  For the second part of EX1, $\vec{X} = \vec{x}$
would have to be such that, with high probability, setting $\vec{X}$
to $\vec{x}$ would result in an output of ``no tumor''.  For the first 
part of EX1, we need to find a minimal subset of pixels that includes
a pixel in $\vec{X}$ such that changing the values of these pixels
would result in a label of ``tumor''.

There is a subtlety here though.  A tumor is a rare event.
Overall, the probability that someone develops a brain tumor in
their lifetime is less than $1\%$, and the probability 
that a random person on the street has a brain tumor at this moment is
much lower than that.  Suppose that the classifier derived its probability
using MRI images from a typical population.  Given an image $I$ that
is (correctly) labeled ``no tumor'', let $X$ be a single pixel whose
value in $I$ is $x$ such that $X$ is part of an explanation $\vec{X} =
\vec{x}$ of the
label ``tumor'' in a different image $I'$, and in $I'$, $X = x' \ne x$.
Then $X=x$ is an explanation
of ``no tumor'' with goodness $(\alpha,\beta)$ for quite hight
$\alpha$ and $\beta$: in most images where $X=x$, the output is ``no
tumor'' (because the output is ``no tumor'' with overwhelming
probability), and changing $X$ to $x'$, as well as the values of other
pixels in
       $\vec{X}$, we typically get an output of ``tumor''.  But
$X=x$ does not seem like a good explanation of why we believe there is
no tumor!

To deal with this problem (which arises whenever we are classifying a
rare event), we (a) assume that the probability is derived from
training on MRI images of patients who doctors suspect of having a
tumor; thus, the probability of ``tumor'' would be significantly
higher than it is in a typical sample of images; and (b) we would expect
an explanation of ``no tumor'' to have goodness $(\alpha,\beta)$ for
$\alpha$ and $\beta$ very close to 1.  With these requirements, an
explanation $\vec{X} = \vec{x}$ of ``no tumor'' would include pixels
from all the most likely tumor sites, and these pixels would have values
that would allow us to preclude there being a tumor at those sites
(with high probability).
%joe6
This is an instance of a situation where $\K$ would not consist
of all contexts. 

The bottom line here is that we do not have to change the definitions
to accommodate explanations of absence and rare events, although we
have to modify the probability distribution that we consider.
%joe4*: added
That said, finding an explanation for ``no tumor'' seems more
complicated than finding an explanation for ``tumor''.  The standard
approaches that have been used do not seem to work. 
%hana3
In fact, none of the existing image classifiers is able to output
explanations of 
%hana4* only one reason
absence. The reason for this is that, due to the 
%%joe6
%%absence. The reason is two-fold. First, due to the intractability of
%absence. There seem to be two reasons for this. First, due to the
intractability of 
the exact 
computation of explanations, all existing black-box image classifiers construct
some sort of \emph{ranking} of pixels of an input image, which is then used
%hana4*
as a (partially) ordered list from which an approximate explanation is constructed greedily.
%as an explanation or as a basis for constructing an approximate explanation.
%joe6*: So why does this first reason mean that they can't deal with
%explanations of absence?
%hana4* there is only one reason - the fact that ranking doesn't work. not sure
%why we ended with two :)
%hana4
Unfortunately, for explaining absence, there is no obvious ranking of pixels
%%joe6
%%Second, for the negative classification there is no obvious ranking of pixels
%Second, for an explain of absence, there is no obvious ranking of pixels
of the image: since a brain tumor can appear in any part of the brain, all
pixels are equally important for the negative classification.
%joe6*: I don't understand the next sentence.  Are you arguing that
%explanations of absence would be pure noise?  I don't see why.  I cut
%the sentence for now.
%Due to the ranking
%methods being approximate algorithms (as the ranking problem itself is also
%intractable), there are some differences in the ranking, and
%explanations constructed 
%from the ranking are purely noise.

%joe6*: I don't think that anyone would seriously make this argument; rewrote
%One can argue that explanations of absence are not needed, exactly
%because it is not  clear how to explain why something is not there.
%Consider, however, the following scenario: 
%an AI classifier outputs ``no tumor'' on an input image, yet the
In general, people find it difficult to explain absences.
%hana4 is it one "do" or two?
%joe11: there should have been two; they can do it and the do do it.
%Nevertheless, they can and do this.
Nevertheless, they can and do do it.
%Nevertheless, they can and do do this.
%
%Suppose that
%an AI classifier outputs ``no tumor'' for an input image, yet the
%radiologist believes there 
%joe6
%is a tumor. What kind of information can the classifier supply with
%the classification to 
%explain its ``reasoning'' to the radiologist?
%joe6
%It seems that if it was a discussion between two 
%human experts, one could explain their decision of ``no tumor'' by
%pointing out the most suspicious
For example, an expert radiologist might explain 
their decision of ``no tumor'' to another expert by
pointing out the most suspicious 
%joe6*
%region of the brain. But this is exactly what Definition~\ref{def:PEX}
region(s) of the brain and explaining why they did not think there was
a tumor there, by indicating why the suspicious features did in fact
not indicate a tumor.
But this is exactly what Definition~\ref{def:PEX}
provides, for appropriate values 
of the probabilistic bounds $(\alpha,\beta)$.
%joe6*: rewrote
%That is, for most suspicious regions, the explanation will include
%pixels that cannot be a part 
%of a tumor, which hopefully counts as a sufficient evidence.  
%If the expert does not find what the algorithm considers to be
%sufficient evidence against a tumor sufficiently persuasive, there's a
%problem somewhere else -- either with the algorithm or with the expert.
%joe15
%First, note that we can get an explanation to focus on the most

Indeed, note that we can get an explanation to focus on the most
suspicious regions by making $\beta$ sufficiently small.  Since
explanations must be minimal, the explanation will then return a
smallest set of regions that has total probability (at least) $\beta$.%
\footnote{There will be may be many choices for this; we can 
add code to get the choice that involves the smallest set of
pixels.  These will be the ones of highest probability.}
Alternatively, we can just restrict $\K$ to the most suspicious
regions, by considering only contexts where nonsuspicious regions have
all their pixels set to white, or some other neutral color (this is
%joe8
%yet another advantage of considering $\K$ rather than all of $\U$).
yet another advantage of considering $\K$ rather than all of $\R(\U)$).  
%joe15
%Second, to explain why there is no tumor in a particular region, 
In addition, to explain why there is no tumor in a particular region, 
the expert would likely focus on certain pixels and
say ``there's no way that those pixels can form part of a tumor''; that's
exactly what the ``sufficiency'' part of the explanation does.   The
expert might also point out pixels that would have to be different in
order for there to be a tumor; that's exactly what the 
``necessity'' part of the explanation does.  Of course, this still
must be done for a number of regions, where the number is controlled
by $\beta$ or the choice of $\K$.  Thus, it seems to us that the
definition really is  
doing a reasonable job of
providing an explanation in the spirit of what an expert would provide.

%joe15: moved up, and rewrote slightly2
We can get simpler (although perhaps not as natural) explanations for
the absence of tumors by taking advantage of domain knowledge.  
For example, if we know the minimal 
size of tumors, explaining their absence
can be done by providing a ``net'' of pixels that cannot be a part
of a brain tumor, in which 
the distance between the neighboring pixels is smaller than the size
of a minimal tumor.

\bigskip

%hana3
\commentout{
%hana2
Indeed, consider black-box approaches that compute a ranking of pixels
and then translate it either to 
a heatmap (SHAP) or to a highly ranked area of the image that is sufficient for the classification 
(LIME and ReX). What is the ranking of pixels with respect to the ``no tumor'' classification? Since a
brain tumor can appear in any part of the brain, all pixels are
equally important for the negative classification.
%joe4: I didn't understand the next sentence (or why you need it)
Differences in ranking are due to the methods being approximate, and not precise algorithms, hence the
explanations constructed from the ranking are purely noise.
%hana2 will check about gradcam
%hana2 will add illustrations

 [[HANA, WE NEED
    SOME DISCUSSION OF WHY THE STANDARD APPROACHES DON'T WORK,
    INCLUDING LIME AND SHAP.  WE COULD THEN MAYBE GIVE YOUR APPROACH,
    BUT THEN WE WOULD ALSO PROBABLY WANT EXPERIMENTS SHOWING THAT IT
    DOES WORK REASONABLY WELL.]]
} %end of commented out text

%\section{Explaining Rare Phenomena}

%joe1: folded this into the previous section
\commentout{
The definitions of explanations of image classifiers' outputs in
Section~\ref{sec:imagexp} work according to our intuition in cases
where the   
labels are not rare. However, in many scenarios where explanation
tools are used, important phenomena might be quite rare. Overall, the
probability that a person will develop a brain tumor in their lifetime
is less than $1\%$, and the probability 
that a random person on the street has a brain tumor at this moment is
much lower than that. Since the answers of image classifiers are 
only correct with a certain probability, it seems that a classifier
that always answers ``no tumor'' actually has a good accuracy. This,
however, does not 
seem right and definitely would not help diagnostics. 
}

%joe11:
We conclude by repeating the point that we made in the introduction (now
with perhaps more evidence): while the analysis of MMTS
shows that a simplification of Halpern's definition can go a long way
to helping us understand notions of explanation used in the
literature, we can go much further by using Halpern's actual
definition, while still retaining the benefits of the MMTS analysis.
However, there is still more to be done.  Moreover, dealing with the
full definition may involve added computational difficulties.  We
believe that using domain knowledge may well make things more
tractable, although this too will need to checked.  For example, 
%joe11*: Hana, can we give examples of how domain knowledge might help?
%hana7*
%joe15
%if we are trying to explain the absence of brain tumors at MRI scans,
%and we know the minimal 
%size of tumors, explaining their absence
%boils down to providing a ``net'' of pixels that cannot be a part
%of a brain tumor, in which 
%the distance between the neighboring pixels is smaller than the size
%of a minimal 
%tumor.
We showed above how domain knowledge could be used to provide simpler
explanations of the absence of tumors.
For another example, if we are explaining the absence of cats
in an image of a seascape, 
%joe15
%the
knowledge of zoology allows to eliminate the sea part of the image
as a possible area where cats can be located, as cats are not marine animals.
%joe12:
This seems doable in practice.

%joe6: added; please check that your funding support is accurate!

%hana13 newpage instead of a bigskip
%%joe11: added
%\bigskip
\newpage

%hana11 commented out for submission - uncomment for the CR
%\noindent {\bf Acknowledgments:} We thank Ilse van der Linden for
%pointing us to the work of MMTS.
%Hana Chockler was supported in part by the UKRI Trust-worthy Autonomous Systems Hub (EP/V00784X/1)
%and the UKRI Strategic Priorities Fund to the UKRI Research Node on Trustworthy Autonomous Systems Governance and Regulation (EP/V026607/1).
%Joe Halpern was supported in part by
%NSF grant CCF 2319186,
%ARO grant W911NF-22-1-0061, MURI grant W911NF-19-1-0217, and a grant
%from the Cooperative AI Foundation.

%joe6: I moved one reference from expai to joe.bib, and removed
%unneeded bib refs
%\bibliography{hc,joe,z,aaai24,expai}
%joe6
%hana11 IJCAI style
\bibliographystyle{named}
%\bibliographystyle{chicago}
%joe9: I moved the references to joe.bib
%\bibliography{joe,z,expai}
\bibliography{joe,z}

\end{document}